\documentclass[runningheads]{llncs}

\usepackage[utf8]{inputenc}
\usepackage[pdftex]{graphicx}
\usepackage{amsmath}
\usepackage{amssymb}
\usepackage{booktabs}
\usepackage[dvinames]{xcolor}

\usepackage{hyperref}
\hypersetup{pdfborder = {0 0 0}}

\begin{document}

\title{Alzheimer's Disease Diagnosis via Deep Factorization Machine Models}
\author{
  Raphael Ronge\inst{1} \and
  Kwangsik Nho\inst{2} \and
  Christian Wachinger\inst{1} \and
  Sebastian P{\"{o}}lsterl\inst{1}} %
% index{Ronge, Raphael}
% index{Nho, Kwangsik}
% index{Wachinger, Christian}
% index{P{\"{o}}lsterl, Sebastian}
\authorrunning{R.~Ronge et al.} %
\institute{Artificial Intelligence in Medical Imaging (AI-Med),\\
    Department of Child and Adolescent Psychiatry,\\
    Ludwig-Maximilians-Universit{\"{a}}t, Munich, Germany \\
\and
Department of Radiology and Imaging Sciences, and \\
the Indiana Alzheimer's Disease Research Center, \\
Indiana University School of Medicine, Indianapolis, IN, USA
} %
\maketitle              %

\begin{abstract}
The current state-of-the-art deep neural networks (DNNs)
for Alzheimer's Disease diagnosis
use different biomarker combinations to classify
patients, but do not allow extracting knowledge about
the interactions of biomarkers.
However, to improve our understanding of the disease, it is
paramount to extract such knowledge from the learned model.
In this paper, we propose a Deep Factorization Machine model that combines the ability of DNNs
to learn complex relationships and the ease of
interpretability of a linear model.
The proposed model has three parts:
(i) an embedding layer to deal with sparse categorical data,
(ii) a Factorization Machine to efficiently learn
pairwise interactions,
and (iii) a DNN to implicitly model
higher order interactions.
In our experiments on data from the  Alzheimer's Disease Neuroimaging Initiative, we demonstrate that
our proposed model classifies
cognitive normal, mild cognitive impaired, and
demented patients more accurately than competing models.
In addition, we show that valuable knowledge about
the interactions among biomarkers can be obtained.

\end{abstract}

\section{Introduction}\label{sec:Introduction}
    Alzheimer's Disease (AD) patients account for 60--80\% of all dementia cases~\cite{Fan2020}.
    Worldwide, 50 million patients have dementia and their number is estimated to triple by 2050~\cite{patterson2018world}.
    AD is a neurodegenerative disease whose progression is highly heterogeneous and not yet fully understood~\cite{Scheltens2016}.
    Mild cognitive impairment (MCI) is a pre-dementia stage which
    results in cognitive decline, but not to an extent that it
    impairs patients' daily live~\cite{Petersen2011}.
    Subjects with MCI are at an increased risk of
    developing dementia due to AD, which would make them
    completely dependent upon caregivers~\cite{Petersen2011}.
    This transition is complex and not yet fully understood.
    Therefore, research in the last decade focused on identifying biomarkers
    to infer which stage of the disease a patient is in~\cite{Jack2013}.
    Important biomarkers include demographics,
    brain atrophy measured by
    magnetic resonance images (MRI), and
    predispositions due to genetic alterations in the form
    of single nucleotide polymorphisms~(SNPs)
    (see \cite{Scheltens2016} for a detailed overview).

    For accurate patient stratification it is important to
    also consider the inter-relationships between biomarkers
    and model their interactions.
    Deep learning techniques excel at implicitly learning complex
    interactions, but extracting this knowledge is
    challenging due to their black-box nature~\cite{BarredoArrieta2020}.
    At the other end of the spectrum are linear models that
    are highly interpretable, but
    only account for interactions when those are explicitly specified.
    Hence, approaches that can model complex interactions while
    preserving interpretability are required to further
    improve our understanding of the interaction
    between biomarkers.

    In this work, we propose a model that is able to
    utilize both low- and high-order feature interactions.
    Our model comprises two parts: (i) a Factorization Machine (FM) that
    explicitly learns pairwise feature interactions without the need of
    feature engineering,
    and (ii) a deep neural network (DNN) that can learn arbitrary low- and
    high-order feature interactions implicitly.
    Consequently, our model preserves the best of both worlds:
    the interpretability of linear models -- via the FM --
    and the discriminatory power of DNNs.
    In our experiments, we demonstrate that our proposed model
    outperforms competing methods for
    classifying healthy controls, patients with MCI,
    and patients with AD.

\section{Related Work}\label{sec:RelatedWork}

    Several existing works study the fusion of multi-modal
    data for AD diagnosis and the interaction between features.
    Zhang~et~al.~\cite{Zhang2011} use MRI, FDG-PET,
    and biomarkers derived from cerebrospinal fluid
    (CSF). %
    They use a multiple-kernel SVM that uses one
    kernel per modality and combines modalities by a weighted
    sum of modality-specific kernels.
    This way, interactions can only be addressed implicitly
    by absorbing them into the sum of kernels and interpretability
    is lost.
    Tong~et~al.~\cite{Tong2017} introduce a non-linear graph fusion
    approach for multi-modal AD diagnosis.
    Their approach can assign
    an overall importance value to each modality in a manner that scales
    independently of the number of features.
    However, this does not allow for modelling interactions
    between single features.
    Khatri~et~al.~\cite{Khatri2020} use an Extreme Learning Machine (ELM) -- a single layer feed-forward neural network (NN).
    They use regional volume and thickness measurements, CSF biomarkers %
    ApoE allele information, and the Mini-Mental State Examination (MMSE) cognitive score.
    Because of the use of MMSE, their model is not solely based on biological measurements, but includes diagnostic information, which usually is among the variables of interest.
    Moreover, ELMs are a type of neural network, which makes their interpretation difficult~\cite{Olden2004,Tsang2017}.
    In~\cite{Venugopalan2021}, a ``Multimodal deep learning [model] for early detection of Alzheimer's Disease stage'' is proposed.
    They account for the lack of interpretability of DNNs, by running the model multiple times with one feature masked at a time. The sharper the drop in performance, the higher they rank the importance of the masked feature. While this provides a measure of importance on a per-feature level, it ignores how the model utilizes feature interactions.
    Ning~et~al.~\cite{Ning2018} use a two hidden layer NN to perform AD diagnosis based on MRI-derived features and genetics.
    They attempt to determine the importance of features and feature interactions via back-propagation based on the partial derivatives method~\cite{Gevrey2003}. The authors point out that
    it remains to be tested how well the computed importance measure reflects the actual prediction computation by their model.

    Finally, we want to emphasize that all the existing
    approaches only study feature interactions by
    trying to disentangle what the model learned in a post-hoc
    manner, but not how changes to the model architecture can
    make the model itself more interpretable.

    \begin{figure}[t]
        \centering
        \includegraphics[width=0.7\textwidth]{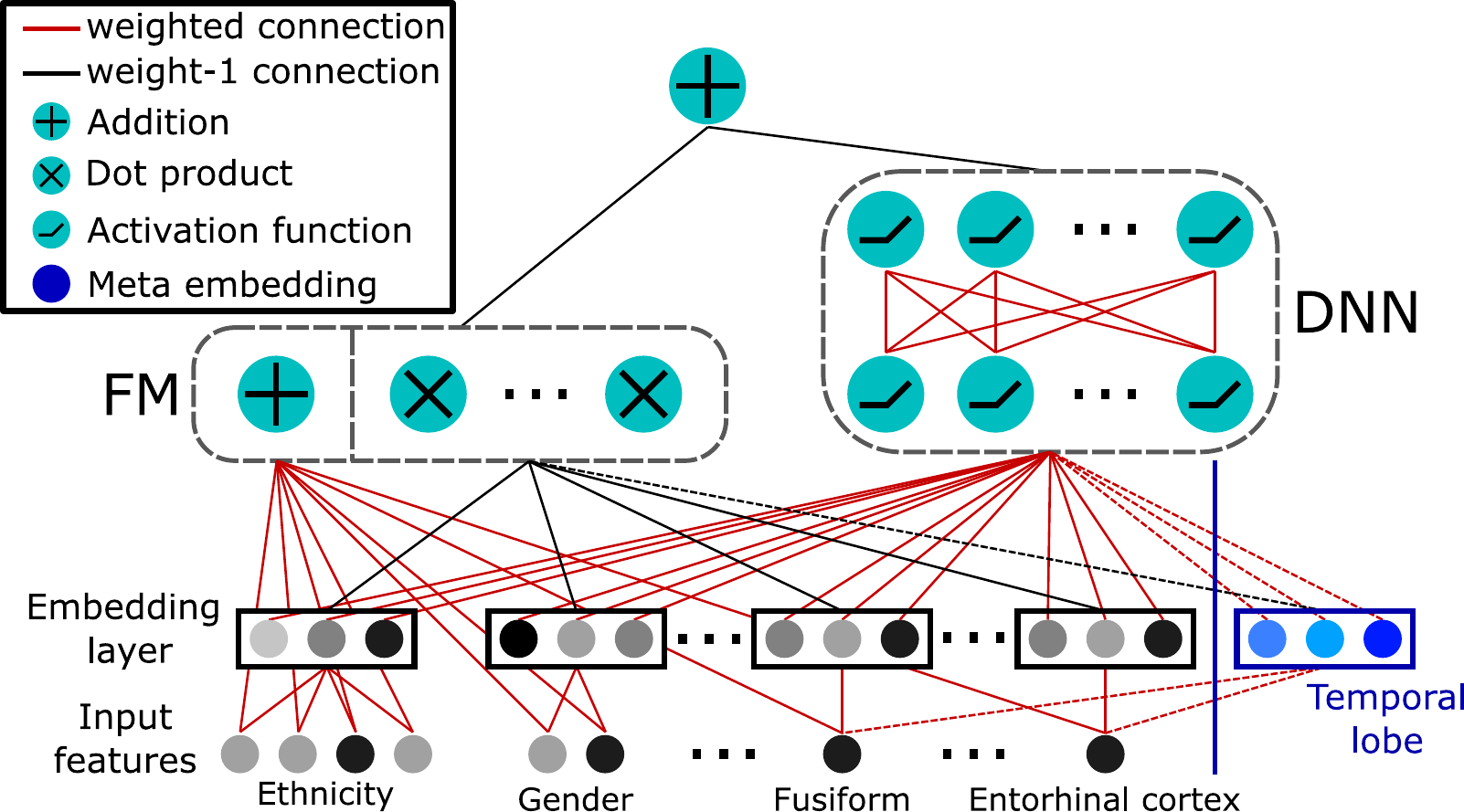}
        \caption{Overview of the proposed model.}\label{fig:DeepFMModelStructure}
    \end{figure}

\section{Methods}\label{sec:Methods}
    The proposed model comprises three major parts
    for improved AD diagnosis (see fig.~\ref{fig:DeepFMModelStructure}).
    The first part is an embedding layer to deal with sparse data~\cite{Guo2017}.
    The second part is based on the Factorization Machine~\cite{Rendle2010}, which models pairwise feature interactions as an inner product of latent vectors from the
    embedding layer.
    The third part is a Deep Neural Network~(DNN) that
    has the potential to implicitly learn complex feature interactions.
    The combined model is closely related to the
    DeepFM~\cite{Guo2017} for click-through-rate prediction.
    We train our model to differentiate between three groups: AD, MCI and CN patients.

    \subsection{Embedding Layer}\label{subsec:EmbeddingLayer}

        The first layer is an embedding layer, similar to the one in~\cite{Guo2017}.
        The layer serves two purposes: First, DNNs are unable to train on sparse data, and second, one feature can span multiple columns if it is e.g. a one-hot encoded categorical feature.
        Via the embedding, each feature is represented as one dense vector, which leads to a more comprehensible representation.
        The embedding layer condenses each $d$-dimensional feature
        $\mathbf{x}_i$ into a vector $\mathbf{e}_i$ with fixed length $m$: $\mathbf{e}_i = \mathbf{A}_i \mathbf{x}_i$,
        where $\mathbf{A}_i \in \mathbb{R}^{m\times d}$ is the learned embedding matrix for feature $i \in \{ 1,\ldots,n \}$.
        The embedding vector $\mathbf{e}_i$ represents the entire
        feature and eliminates the problems that arise when
        training neural networks on sparse data --
        in particular,
        when categorical features with many
        categories are present,
        because the embedding layer reduces the dimensionality
        compared to a one-hot encoding.

        Another advantage of the embedding layer is that
        it can be used to combine sets of features
        describing whole brain areas.
        For instance, one can combine the volume measurements
        of all regions belonging to the temporal lobe into
        one embedding vector.
        Therefore, the embedding layer can be used as a mean to incorporate
        domain knowledge about the structural or functional
        relationship between features.
        As AD is a highly heterogeneous disease and brain regions
        are strongly interrelated, this can improve
        predictive performance as well as interpretability
        of interaction effects, which we will discuss next.

    \subsection{Factorization Machine}\label{subsec:FactorizationMachines}
        The Factorization Machine~(FM; \cite{Rendle2010})
        consists of three parts: a bias,
        a linear predictor, and a pairwise-interaction term
        (for simplicity, we omit the bias in fig.~\ref{fig:DeepFMModelStructure}).
        For an $n$-dimensional feature vector $\mathbf{x}$,
        the FM for class $c$ is defined as:
        \begin{equation}\label{equ:FM}
            \hat{y}_\text{FM}^c(\mathbf{x}) =
            \underbrace{w_0}_\text{bias} +
            \underbrace{\textstyle\sum^{n}_{i=1} w_i x_i}_\text{linear predictor} +
            \underbrace{\textstyle\sum^{n}_{i=1}\sum^{n}_{j=i+1} \langle \mathbf{e}_i, \mathbf{e}_j \rangle x_i x_j}_\text{interaction term}
        \end{equation}
        The key idea of the FM is to not learn
        interaction weights explicitly, which would scale
        quadratically in the number of features,
        but implicitly through the dot product
        $\langle \mathbf{e}_i, \mathbf{e}_j \rangle$.
        Hence, weights are
        shared across interaction terms and
        one has to learn $n$ embedding matrices $\mathbf{A}_i$ instead of $n^2$ weights that would be required
        for explicit interaction modelling.
        To preserve the linear complexity of
        the linear model, while still accounting
        for pairwise interactions, we
        reformulate the pairwise-interaction computation as in~\cite{Rendle2010},
        resulting in a $\mathcal{O}(kn)$ runtime.

    \subsection{Deep Factorization Machine}\label{subsec:DeepFactorizationMachines}
        So far, our model only accounts
        for linear and pairwise interactions.
        We account for high-order interactions implicitly by
        employing a DNN alongside
        the FM from above~\cite{Guo2017} (see fig.~\ref{fig:DeepFMModelStructure}).
        The DNN receives the concatenated
        embedding vectors $\mathbf{e}_i$, and thus is
        equipped to learn from high-dimensional sparse data.
        The DNN contains two hidden layers with ReLU activation function $\sigma(x) = \max(0,x)$:
        \begin{equation}
            \hat{y}_\text{DNN}^c(\mathbf{x}) =
            \sigma ( \mathbf{W}^{(1)} \cdot
            \sigma (
                \mathbf{W}^{(0)} \cdot
                \mathrm{CONCAT} (\mathbf{e}_1, \ldots, \mathbf{e}_n )
                + \mathbf{b}^{(0)}
            )
            + \mathbf{b}^{(1)}
            ) ,
        \end{equation}
        where $\mathbf{W}^{(k)}$ and $\mathbf{b}^{(k)}$ are
        the weight matrix and bias of the $k$-th layer, respectively.
        Finally, the overall prediction for class $c$ of our DeepFM model is:
        \begin{equation}
            \hat{y}^c(\mathbf{x}) = \mathrm{Softmax}\left(
            \hat{y}_\text{FM}^c(\mathbf{x}) + \hat{y}_\text{DNN}^c(\mathbf{x})
            \right) ,
        \end{equation}
        where $\hat{y}_\text{FM}^c$ is the factorization machine
        defined in equation~\eqref{equ:FM}.
        During training, we optimize the weights of the FM part
        ($w_0,\ldots,w_n$), the embedding matrices $\mathbf{A}_1,\ldots,\mathbf{A}_n$,
        which are shared among the FM and deep part of our model,
        and the parameters of the DNN
        ($\mathbf{W}^{(0)}, \mathbf{W}^{(1)}, \mathbf{b}^{(0)}, \mathbf{b}^{(1)}$).

\section{Experiments}\label{sec:Experiments} %
    We evaluated the proposed model on data provided by the Alzheimer's Disease Neuroimaging Initiative~\cite{Jack2008}.
    Table~\ref{tab:Demographics} summarizes the data.
    Our dataset contains a total of 1492 patients with 6844 visits and
    three class labels: AD (1536 visits), MCI (3131 visits), cognitive normal (CN; 2177 visits).
    In addition to demographic data, we collected for each patient MRIs and processed them with
    FreeSurfer~\cite{Fischl2012} to obtain 20 volume and 34
    thickness measurements.
    Moreover, we collected Amyloid-$\beta$ (A$\beta$), Tau, and phosphorylated Tau (pTau) concentration in CSF.
    Finally, we collected 41 genetic markers, previously shown to be associated with
    AD and atrophy~\cite{Hibar2015,Lambert2013}, as described in~\cite{Wachinger2018}. Except for CSF measurements, which are only available for 1863 visits, each modality is available for all patients. A$\beta$, Tau, and pTau are important biomarkers and in order to not heavily reduce the available data, we keep them and handle missing values as zero, to get as good as a prediction as possible.
    In total, we used 109 features.

    \begin{table}[t]
    \centering
    \scriptsize
    \caption{Overview of the data used in our experiments
    (MMSE is not used as feature).}\label{tab:Demographics}
    \begin{tabular}{l p{8mm} p{9mm} p{9mm} p{9mm} p{9mm} p{9mm} p{9mm} p{9mm} p{9mm}} %
        \toprule
        Feature             & \multicolumn{3}{l}{AD-patients}               &  \multicolumn{3}{l}{MCI-patients}             & \multicolumn{3}{l}{NC-patients}           \\
                            & min   & mean  & max                           &  min  & mean  & max                           & min   & mean  & max                       \\ \midrule
        MMSE     & 10.0  & 21.9  & 30.0                          & 10.0  & 27.5  & 30.0                          & 20.0  & 29    & 30.0                      \\
        Age                 & 55.0  & 74.4  & 90.9                          & 54.4  & 73.1  & 91.4                          & 55.0  & 73.9  & 90.1                      \\
        Education (Years)   & 4.0   & 15.45 & 20.0                          & 6.0   & 16.03 & 20.0                          & 6.0   & 16.43 & 20.0                      \\
        Gender              & \multicolumn{3}{l}{58.3 \% Male}              & \multicolumn{3}{l}{61.8\% Male}               & \multicolumn{3}{l}{52.3 \% Male}          \\
        \bottomrule
    \end{tabular}
    \end{table}

    To avoid data leakage due to confounding effects of age and sex~\cite{Wen2020},
    we split the data into 5 non-overlapping folds using only baseline visits
    such that diagnosis, age and sex are balanced across folds~\cite{Ho2007}.
    We used one fold as test set and combined the remaining folds such that
    80\% of it comprise the training set and 20\% the validation set.
    We extended the training set, but not validation or test, by including each patient's
    longitudinal data.

    We optimized the models' hyperparameters (see table~\ref{tab:Hyperparameters}) for each of the five folds separately
    via Bayesian black-box optimization
    on the validation set~\cite{head_tim_2020_4014775}.
    We compare the proposed DeepFM to a standalone DNN,
    the FM~\cite{Rendle2010}, and
    a linear logistic regression model that explicitly
    accounts for all pairwise interactions.
    Each model is evaluated by the balanced accuracy on the respective test set.
    \begin{table}[ht]
        \centering
        \scriptsize
        \caption{Hyperparameter Search Space. $\mathcal{U}$ uniform-/$\mathcal{LU}$ log-uniform-distribution.}\label{tab:Hyperparameters}
        \begin{tabular}{lllll}
            \toprule
            Hyperparameter              & DeepFM                            & FM                                & DNN                               & Linear          \\\midrule
            Neurons Layer 1             & $\mathcal{U}(1, 400)$             & -                                 & $\mathcal{U}(1, 400)$             & -                             \\
            Neurons Layer 2             & $\mathcal{U}(1, 400)$             & -                                 & $\mathcal{U}(1, 400)$             & -                             \\
            Neurons Layer 3             & -                                 & -                                 & $\mathcal{U}(0, 400)$             & -                             \\
            Length Embedding Vector    & $\mathcal{U}(1, 20)$              & $\mathcal{U}(1, 20)$              & $\mathcal{U}(1, 20)$              & -                             \\
            Learning Rate               & $\mathcal{LU}(1e^{-04}, 0.9)$     & $\mathcal{LU}(1e^{-04}, 0.9)$     & $\mathcal{LU}(1e^{-04}, 0.9)$     & $\mathcal{LU}(1e^{-04}, 0.9)$ \\
            L1-Regu. Weight    & $\mathcal{LU}(1e^{-04}, 0.9)$     & $\mathcal{LU}(1e^{-04}, 0.9)$     & $\mathcal{LU}(1e^{-04}, 0.9)$     & $\mathcal{LU}(1e^{-04}, 9)$   \\
            L2-Regu. Weight    & $\mathcal{LU}(1e^{-04}, 0.9)$     & $\mathcal{LU}(1e^{-04}, 0.9)$     & $\mathcal{LU}(1e^{-04}, 0.9)$     & $\mathcal{LU}(1e^{-04}, 9)$   \\
            Dropout                     & $\mathcal{LU}(0.1, 0.9)$          & -                                 & $\mathcal{LU}(0.1, 0.9)$          & -                             \\
            \bottomrule
        \end{tabular}
    \end{table}

\section{Results \& Discussion}\label{sec:ResultsDiscussion}

\begin{figure}[tb]
        \centering
        \includegraphics[scale=0.2]{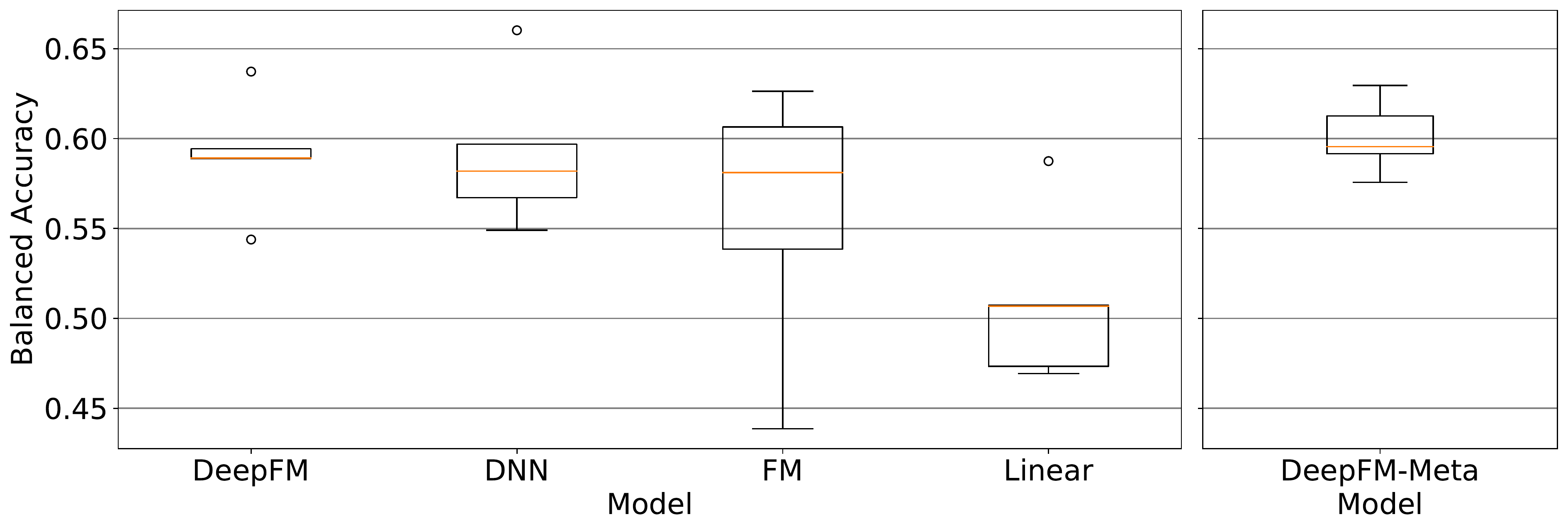}
        \caption{Balanced Accuracy Comparison}
        \label{fig:BalancedAcc}
    \end{figure}

    \paragraph{Performance.}
    With a median balanced accuracy of 0.589, DeepFM has the highest performance of all models (see fig.~\ref{fig:BalancedAcc}). It is slightly better than DNN ($\text{acc}_{med} = 0.582$) and FM ($\text{acc}_{med} = 0.581$). DeepFM improves over the DNN thanks to the added FM model. Solely the linear model is unable to achieve similar performance. This shows the effectiveness of the FM approach and its capability to
    learn interactions among many features.
    In addition, we explored meta-embeddings, which combine volume
    measurements of larger brain regions into a single embedding vector.
    We only combine brain volume features as those are the largest feature group and combining them by brain region is medically reasonable.
    Our results demonstrate that this leads to a slightly higher balanced accuracy of 0.596, but most importantly lowers the variance across folds (fig.~\ref{fig:BalancedAcc}, right).
    By combining larger brain regions, the pairwise interaction space shrinks and the model is less prone to overfitting.

    \begin{figure}[tb]
      \centering
      \includegraphics[scale=0.175]{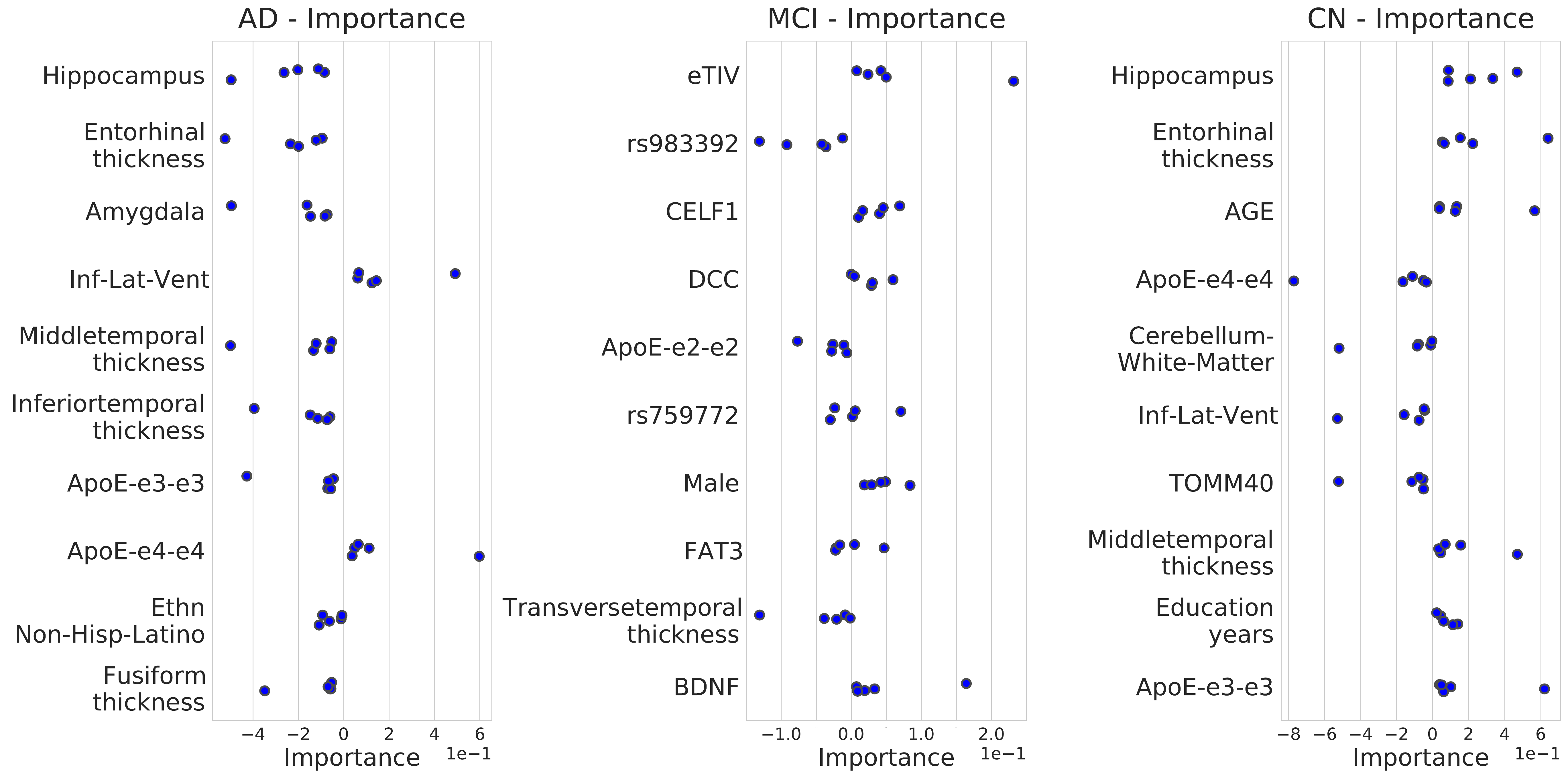}
      \caption{DeepFM - 10 Most Important Linear Features}
      \label{fig:DeepFMImportance}
    \end{figure}
    \begin{figure}[tb]
        \centering
        \includegraphics[scale=0.175]{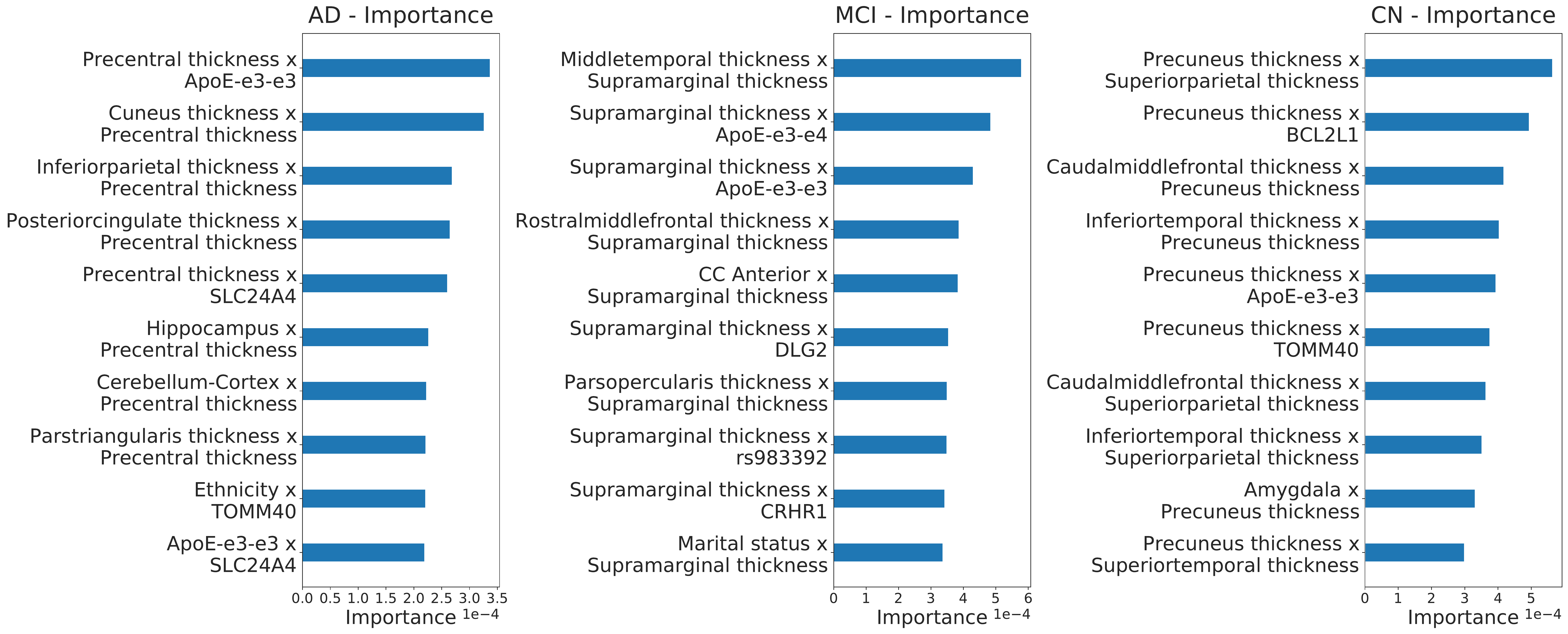}
        \caption{DeepFM - 10 Most Important Feature Interactions}
        \label{fig:DeepFMInteractions}
    \end{figure}
    \begin{figure}[tb]
        \centering
        \includegraphics[scale=0.175]{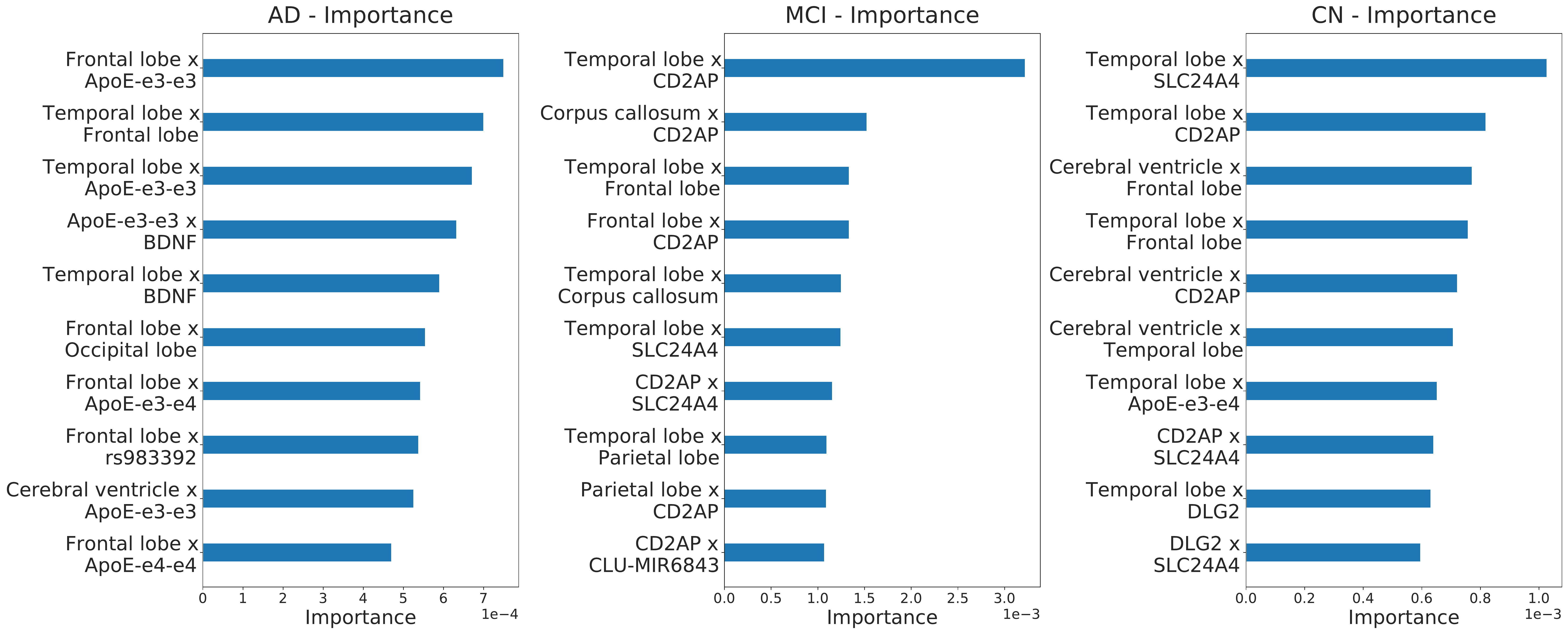}
        \caption{DeepFM-Meta - 10 Most Important Feature Interactions}
        \label{fig:DeepFMMetaInteractions}
    \end{figure}

    \paragraph{Feature Importance.}
        We get a direct measure of feature importance by looking at the weights
        of the linear part of the DeepFM in \eqref{equ:FM}.
        Figure~\ref{fig:DeepFMImportance} displays the weights of the ten most important features
        for each fold (sorted by their mean weight over all folds).
        Because weights are class-specific, we obtain three rankings.
        As the top ten features for the AD (CN) model account for 42\% (39.8\%) of the total feature importance, we only analyze those in more detail.

        The first and third most important features for AD are the volumes of Hippocampus and Amygdala, both lie in the temporal lobe, a region typically affected by AD~\cite{Scott1991,Teipel2006}. The thickness measures in the top features for AD lie in the parietal and temporal lobe. Both regions were previously shown to exhibit atrophy in AD patients~\cite{Dickerson2008,Singh2006}.
        The volume of the lateral ventricle is dependent on the atrophy of the brain regions surrounding it.
        Two variants of the genetic marker ApoE, e3-e3 and e4-e4, are important for prediction, which is reassuring because it
        is an important marker for AD~\cite{Reiman2019,Saykin2010}.
        Being not Hispanic, or not Latino has a slightly positive influence on the AD risk, which is supported by previous studies~\cite{tang2001incidence,Tang1998}.

        It is striking that the learned impact of all the top features for AD and CN
        prediction
        is in line with medical findings.
        Consequently, the weights' signs for the prediction of CN have the opposite sign
        compared to those for AD.
        For MCI, the picture is less clear.
        As MCI patients are part of a complex transition, this group is more heterogeneous.
        While for AD and CN a few very important features are sufficient for prediction, for MCI,
        the top ten features only make up for 23.7\% of the total feature importance and learned weights are incoherent.

    \paragraph{Feature Interactions.}
        The second information that is easily accessible is the importance of feature interactions (see fig.~\ref{fig:DeepFMInteractions}). This gives DeepFM an advantage over DNNs, for which
        extracting information about learned interactions is challenging.
        The information of a feature in an interaction is interconnected with their own and their partners embedding matrix. We extract the importance information by running the model with the test data and computing the relative importance of each single interaction and the output of the rest of the model for every patient. We can then plot the importance value for each interaction as the mean over all patients. The 109 features per patient accumulate to 5886 pairwise interactions. Thus, a single interaction has a relatively small contribution.

        AD affects multiple brain regions at every stage of disease progression. The difficulty of pinpointing AD to a specific region becomes apparent for the pairwise interactions, where multiple brain regions interact. Numerous interactions consist of regions in or near the Hippocampus. Some genetic markers also appear. The interactions between them and different brain regions are an interesting finding that needs to be further explored in the future.
        However, in this setup the brain is scattered in many volumes and small differences or measurement errors become much more pronounced. This makes the interpretation difficult.

        At this point, the advantage of the embedding layer comes into play. While interactions between single small brain regions are hard to interpret, the embedding layer can be used to embed larger regions into a so-called \emph{meta embedding}. The interactions between these larger regions give a better picture of important regions (see fig.~\ref{fig:DeepFMMetaInteractions}). Small deviations have less influence and overfitting is reduced.
        Using meta embedding, DeepFM is especially utilizing interactions between ApoE and temporal or frontal lobe volumes.
        Comparing the most important feature interactions to distinguish AD from the rest, with the ones used for CN patients, it becomes even clearer that DeepFM learns medically explainable features. The CN-model focuses especially on the temporal lobe and cerebral ventricle. Both regions are known to be affected early on in the disease~\cite{Teipel2006,Nestor2008}.

\section{Conclusion}\label{sec:Conclusion}
    We proposed a Deep Factorization Machine model that
    combines the strength of deep neural networks to implicitly
    learn feature interactions and the ease of
    interpretability of a linear model.
    Our experiments on Alzheimer's Disease diagnosis
    demonstrated that the proposed architecture is able
    accurately classify patients than competing methods
    and can reveal valuable insights about the interaction
    between biomarkers.

\subsubsection*{Acknowledgements.}
This research was supported by the Bavarian State Ministry of Science and the Arts and coordinated by the Bavarian Research Institute for Digital Transformation,
and the Federal Ministry of Education and Research in the call for Computational Life Sciences (DeepMentia, 031L0200A).

\bibliographystyle{splncs04}
\bibliography{bibliography-selected}

\end{document}